\pdfoutput=1
\documentclass[11pt,a4paper]{article}

\usepackage[english]{babel}
\usepackage[T1]{fontenc}
\usepackage[utf8]{inputenc}

\usepackage{amsmath,amssymb,amsfonts,mathtools}
\usepackage{geometry}
\usepackage{graphicx}
\usepackage{float}
\usepackage{microtype}

\usepackage{mathptmx}

\graphicspath{{./}}
\title{
Observable Performance Does Not Fully Reflect Adaptive System Organization:\\
A Multi-Level Analysis of Gait Dynamics Under Occlusal Constraint
}

\author{
Jacques Raynal$^{1}$,
Pierre Slangen$^{2}$,
Elsa Raynal$^{3}$,
Jacques Margerit$^{4}$\\[0.5em]
{\small $^{1}$Laboratory of Bioengineering and Nanosciences (LBN), University of Montpellier, France}\\
{\small $^{2}$EuroMov Digital Health in Motion, University of Montpellier, IMT Mines Alès, Alès, France}\\
{\small $^{3}$Certified Sophrologist and Dental Assistant, Sensorimotor Practice, Montpellier, France}\\
{\small $^{4}$Emeritus Professor, University of Montpellier, France}\\[0.5em]
{\small Corresponding author: raynal.cab@gmail.com}
}

\date{}

\makeatletter
\renewcommand{\maketitle}{
  \begin{flushleft}
    {\LARGE\bfseries \@title \par}
    \vspace{0.8em}
    {\normalsize \@author \par}
  \end{flushleft}
  \vspace{1.2em}
}
\makeatother

\begin{document}

\maketitle

\begin{abstract}

  In biomechanical systems, observable performance is often used as a proxy for underlying system organization. However, this assumption implicitly presumes a correspondence between output metrics and internal system states that may not hold in adaptive systems.
  
  In this study, the vertical dimension of occlusion (VDO) is considered as a constraint applied to an adaptive neuromechanical system, enabling the exploration of system-level responses under controlled variations. A single-case design in a patient with Parkinson's disease allows an intra-individual analysis across repeated conditions.
  
  The analysis is structured across three complementary levels: (i) aggregated linear metrics describing observable performance, (ii) a conceptual dynamical-systems framework describing temporal organization in state space, and (iii) an exploratory low-dimensional representation of multivariate gait observations obtained through unsupervised embedding.
  
  The revised Level~3 analysis was reconstructed from the original M1 observation-level gait data. The resulting UMAP representation showed substantial overlap among the six occlusal observational probes and did not reveal independently separated condition-specific clusters. OC2.5 and OC3 exhibited a limited displacement of their centroid positions within the selected embedding, but their distributions remained broadly overlapping.
  
  These findings indicate that the selected multivariate representation contains relational information not expressed by the aggregated score, while providing no evidence that the occlusal probes correspond to distinct physiological states.
  A fourth level is proposed as a purely conceptual extension describing potential relationships between system states. This level is not implemented and is not derived from experimental data.
  
  These observations are strictly exploratory and non-causal. The proposed framework does not establish mechanistic, predictive, directional, diagnostic, or clinically prescriptive relationships. The exploratory multivariate representation should be interpreted as a model-dependent visualization of multivariate gait observations rather than as direct evidence of distinct internal physiological states.
  
  \end{abstract}


\section{Introduction}

\begin{figure}[H]
  \centering
  \includegraphics[width=\textwidth]{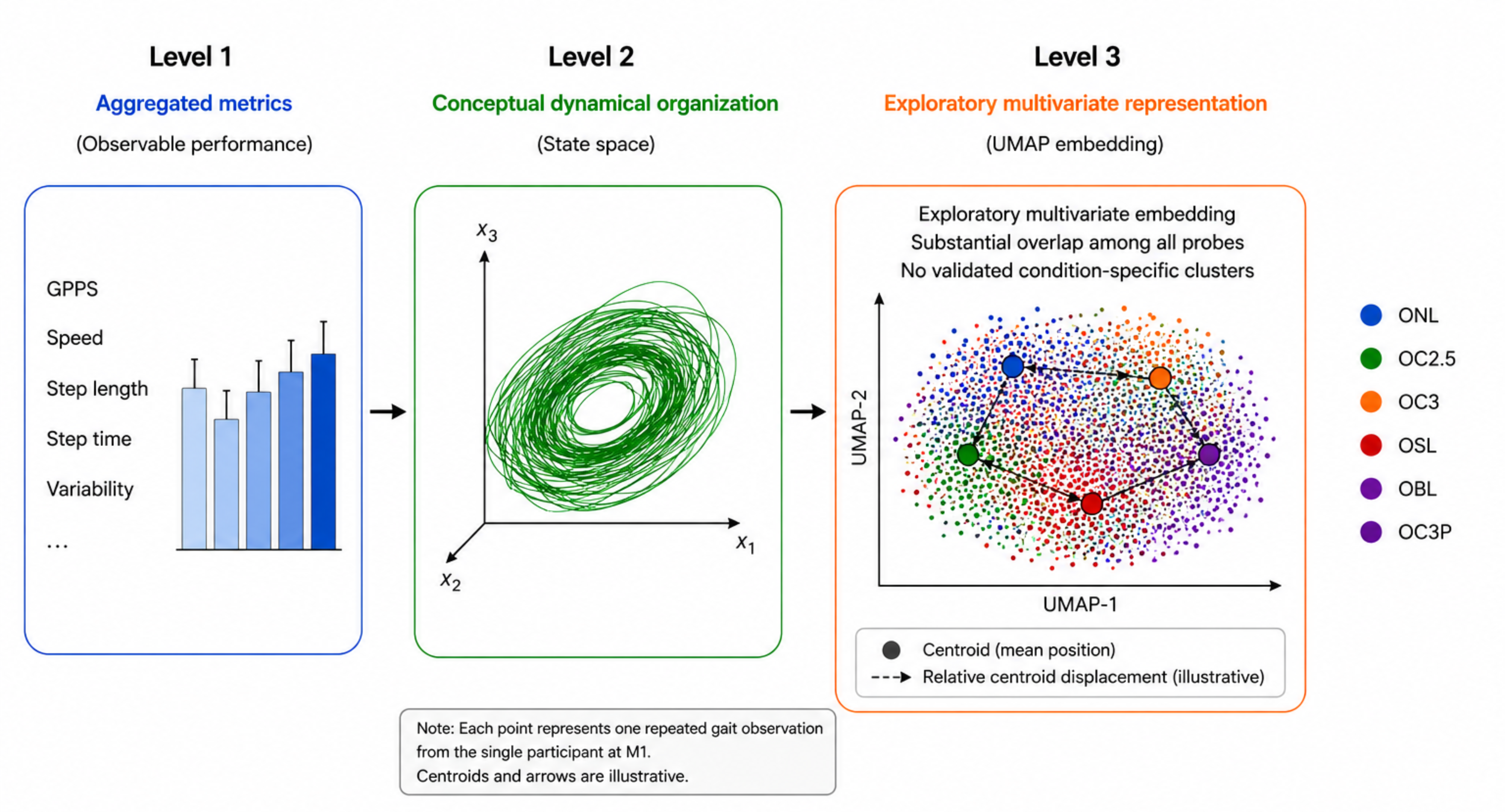}
  \caption{
  Conceptual framework illustrating three complementary levels of gait analysis.
  Level~1 summarizes observable performance using an aggregated scalar representation.
  Level~2 provides a conceptual state-space interpretation of temporal organization.
  Level~3 provides an exploratory low-dimensional embedding of multivariate gait observations.
  The three levels represent complementary analytical views, but none is interpreted as direct access to the underlying physiological state of the system.
  }
  \label{fig:framework}
  \end{figure}

In biomechanical systems, observable performance metrics are commonly used to characterize functional behavior. This approach implicitly assumes that measurable outputs provide sufficient information about the underlying organization of the system. However, in adaptive biological systems, this correspondence may not hold: similar observable performance may arise from different internal system states.
This limitation becomes particularly relevant in contexts where system behavior emerges from interactions between constraints and adaptive mechanisms. In such cases, the identification of a single optimal parameter may not adequately describe system organization. Instead, system behavior must be considered as the result of an interaction between imposed constraints and the adaptive capacity of the system.
The vertical dimension of occlusion (VDO) provides a clinically relevant example of such a constraint. Traditionally described as a measurable anatomical value to be restored, it has often been approached within a static paradigm based on morphological, esthetic, or mechanical references. While these approaches retain clinical utility, their limitations become apparent in complex rehabilitations, where functional instability may occur despite apparently acceptable anatomical reconstruction.
A growing body of observations suggests that VDO cannot be reduced to a fixed intrinsic value specific to the patient \cite{michelotti2011,baldini2013,perinetti2006}. Instead, it may be understood as a constraint applied to an adaptive neuromechanical system, engaging sensorimotor processes over time. Within this perspective, the clinical question shifts from identifying an optimal value to characterizing system behavior under a given constraint.
This reinterpretation is consistent with frameworks from dynamical systems theory, in which global behavior emerges from interactions between system components and constraints \cite{kelso1995,izhikevich2007,peterka2002}. Within this framework, occlusion is not considered as a purely local mechanical phenomenon, but as part of a neurosensory interface integrated into sensorimotor regulation, notably through trigeminal afferents \cite{massion1994,deriu2003,gangloff2002}. Although interactions between occlusion, posture, and locomotion remain debated, available evidence suggests that occlusal conditions may be associated with modulations of global neuromuscular organization \cite{cuccia2009,nowak2023,sakaguchi2007}.
Recent advances in wearable technologies enable the acquisition of gait data under ecological conditions. Spatiotemporal parameters, center of pressure dynamics, and postural indices provide measurable descriptors of system behavior \cite{winter1995,horak2006,ivanenko2018}. However, these aggregated metrics do not capture the full organization of the system, particularly its temporal structure and the relationships between system states.
To address this limitation, we propose a multi-level analytical framework. The first level consists of aggregated metrics describing observable performance. The second level introduces a dynamical-systems perspective, in which gait is interpreted conceptually as the evolution of a system within a state space. The third level relies on an unsupervised low-dimensional embedding of multivariate gait observations \cite{bengio2013,lecun2022,mcinnes2018}. This third level is used as an exploratory visualization of relational structure and not as an independently validated identification of physiological states.
The objective of this study is not to determine an optimal VDO or to demonstrate independently validated condition-specific physiological states. It is to examine how the interpretation of gait observations changes across aggregated and multivariate representations, and to determine whether an unsupervised embedding provides complementary descriptive information beyond a scalar performance score.
This work is intended as an exploratory methodological and conceptual framework for structuring the analysis of constraint-driven biological systems. It is not presented as a predictive, validated, mechanistic, diagnostic, or clinically prescriptive model. The multivariate representations obtained here should not be interpreted as direct evidence of independently validated physiological states.
Figure~\ref{fig:framework} summarizes the proposed analytical architecture.

\section{Materials and Methods}

\subsection{Study Design}

This study aims primarily to characterize system-level organization under constraint, rather than to establish clinical protocols. It is based on a single-case experimental design, enabling an intra-individual analysis of the effects of controlled variations in VDO on gait and postural regulation \cite{smith2012,kazdin2011}.
Two experimental sessions were conducted eleven weeks apart: a baseline condition (M1) and a second session (M2), with identical occlusal conditions tested at both time points. Between M1 and M2, the participant remained under natural occlusion, and no controlled occlusal intervention was introduced.
Within each session, the six occlusal observational probes were assessed in the same fixed order: ONL, OSL, OBL, OC2.5, OC3, and OC3P. The order was reproduced identically at M1 and M2. Because condition order was not randomized, each probe was associated with a defined temporal position within the session; condition-related differences therefore cannot be separated from possible order, adaptation, learning, or fatigue effects.
The study is descriptive and exploratory. It does not aim to isolate or attribute observed effects to a single causal factor.

\subsection{Participant}

The participant was a 72-year-old male diagnosed with Parkinson's disease for 13 years. Gait was autonomous, without the use of assistive devices. The neurological condition is reported to describe the clinical context of the system under study and is not used as an explanatory variable.
Between M1 and M2, the participant underwent eleven individual 45-minute sophrology-oriented sensorimotor sessions aimed at body awareness, postural regulation, and sensorimotor integration. This intervention is reported as part of the longitudinal context but was not controlled or isolated within the study design; no M1--M2 difference can therefore be causally attributed to it \cite{rangelrooij2020}.

\subsection{Occlusal Conditions}
Six occlusal conditions were evaluated:
\begin{itemize}
\item ONL: natural occlusion;
\item OSL: clenched occlusion;
\item OBL: open mouth;
\item OC2.5: increased VDO by 2.5 degrees in centric relation;
\item OC3: increased VDO by 3 degrees in centric relation;
\item OC3P: increased VDO with mandibular protrusion.
\end{itemize}

ONL, OSL, OC2.5, and OC3 preserved the mandibular hinge axis, whereas OBL and OC3P modified mandibular positioning \cite{rugh1981,dawson2007}.
The labels OC2.5 and OC3 refer to the nominal angular settings of the experimental occlusal devices relative to the mandibular hinge-axis construction used in this single case. They should not be interpreted as universally transferable anatomical displacements. OC3P combined the nominal 3-degree increase with mandibular protrusion and therefore represented a distinct compound probe.

\subsection{Gait Acquisition}

Gait data were recorded using instrumented insoles combining pressure sensors and inertial measurement units. Such wearable systems have been used for ecological gait analysis and have shown relevance in neurological and biomechanical contexts \cite{deldin2016,martin2024,mostovoy2023}.

The protocol consisted of a standardized walking path with linear forward and return phases. Only linear walking phases were retained for analysis.

\subsection{Data Processing}

Data were extracted from twelve CSV recordings corresponding to the six probes at M1 and M2. The numbers of recorded observations were: M1--ONL $n=50$, M1--OSL $n=46$, M1--OBL $n=33$, M1--OC2.5 $n=41$, M1--OC3 $n=51$, M1--OC3P $n=49$, M2--ONL $n=60$, M2--OSL $n=58$, M2--OBL $n=57$, M2--OC2.5 $n=60$, M2--OC3 $n=57$, and M2--OC3P $n=62$. These are repeated gait observations from one participant, not independent participants.

Preprocessing was performed in a Python environment. The Level~1 composite-score pipeline and the Level~3 multivariate-embedding pipeline were treated as distinct analytical procedures.

For Level~1, numerical biomechanical variables were screened using the interquartile-range rule, with detected outliers replaced by the median. Technical zeros in derived variables were treated as unavailable values and were imputed according to the documented condition-level median procedure, with a session-level median used only when an entire derived variable was unavailable for one condition.

For the reconstructed Level~3 analysis, the six original M1 files were concatenated before projection. Fifty-five numerical biomechanical variables were retained. Empty technical fields, categorical variables, absolute identifiers, timestamps, and non-informative columns were excluded. Missing numerical values were imputed using the median of the corresponding retained variable within the combined M1 analytical matrix. No IQR-based outlier replacement was applied within the reconstructed Level~3 pipeline. The retained variables were then standardized jointly across the 270 M1 observations before UMAP projection.

All retained variables, imputed values, preprocessing parameters, UMAP coordinates, centroid coordinates, centroid-distance values, and integrity records were archived to permit exact computational reproduction.

The reconstructed Level~3 analysis reported in the present manuscript was performed on the six original M1 recordings only. The M1 dataset comprised 270 repeated gait observations distributed across ONL, OSL, OBL, OC2.5, OC3, and OC3P. Consequently, the Level~3 results describe the baseline session and should not be interpreted as a longitudinal M1--M2 comparison within a shared representation space.

The Level~1 composite representation and the Level~3 multivariate matrix are distinct analytical objects. GPPS is computed from the disclosed nine-variable Level~1 vector, whereas the revised Level~3 embedding is derived independently from the M1 multivariate gait observations without prior GPPS aggregation. The preprocessing operations and dimensionality-reduction parameters used to generate the revised Level~3 figures must be retained with the analytical archive to permit exact computational reproduction.

\subsection{Analytical Framework}
The analysis was structured across three complementary representational levels: aggregated observable performance, conceptual dynamical organization, and exploratory multivariate embedding.

\subsubsection{Level 1: Observable Performance}

Let the conceptual Level~1 variable vector be defined as:

\[
X = (v,c,D,A,A_P,A_L,L,CoP,CAPA),
\]

where $v$ is walking speed, $c$ is cadence, $D$ is step time, $A$ is temporal asymmetry, $A_P$ is single-support asymmetry, $A_L$ is step-length asymmetry, $L$ is step length, $CoP$ represents a center-of-pressure descriptor, and $CAPA$ is a
plantar capacitive index.

Before aggregation, the variables were oriented so that higher values consistently represented better performance. Walking speed, cadence, step length, $CoP$, and $CAPA$ contributed positively, whereas step time and the absolute magnitudes of the asymmetry
variables contributed negatively. Absolute values were used for asymmetry measures because the sign indicates direction rather than severity:

\[
A^{*}=|A|, \qquad
A_P^{*}=|A_P|, \qquad
A_L^{*}=|A_L|.
\]

To preserve comparability across conditions and sessions, each variable was normalized using a common min--max transformation estimated once from the aggregated M1--M2 analytical dataset:

\[
\widetilde{x}
=
\frac{x-x_{\min}^{\mathrm{global}}}
{x_{\max}^{\mathrm{global}}-x_{\min}^{\mathrm{global}}}.
\]

The Level~1 raw global postural performance score was defined as:

\[
\begin{aligned}
GPPS_{\mathrm{raw}}
={}&
0.15\widetilde{v}
+0.15\widetilde{c}
-0.10\widetilde{D}
-0.10\widetilde{|A|}
-0.10\widetilde{|A_P|} \\
&-0.05\widetilde{|A_L|}
+0.10\widetilde{L}
+0.15\widetilde{CoP}
+0.15\widetilde{CAPA}.
\end{aligned}
\]

These coefficients correspond to the historical Level~1 weighting scheme after removal of the longitudinal progression term $I$.
The original historical formulation included a term $+0.10I$. Because $I$ incorporated information related to M1--M2 progression, its inclusion in a score subsequently used to evaluate longitudinal change would introduce circularity. It was therefore excluded from the revised Level~1 representation and reserved conceptually for
longitudinal analysis at Level~4.

The historical formulation also assigned a coefficient of $0.15$ to both $CoP$ and $CAPA$. This weighting was retained in the revised formula. In particular, $CAPA$ was not included with unit weight, because doing so would increase its contribution by a factor of approximately 6.7 relative to the historical specification and could
cause the composite score to be dominated by the plantar capacitive component.

For interpretability, the raw composite score may be transformed to a 1--10 scale using bounds estimated once from the complete analytical dataset:

\[
GPPS
=
1+
9
\frac{
GPPS_{\mathrm{raw}}-GPPS_{\mathrm{raw,min}}
}{
GPPS_{\mathrm{raw,max}}-GPPS_{\mathrm{raw,min}}
}.
\]

The same transformation bounds must be applied to all observations,
conditions, and sessions. Condition-level GPPS values correspond to
the mean of the observation-level scores within each condition and
session.

GPPS is used exclusively as a descriptive Level~1 representation.
It is not presented as a validated, diagnostic, prognostic, or
prescriptive clinical instrument. Its purpose is to test whether an
aggregated representation can discriminate between experimental
conditions. The Level~3 embedding was derived independently from
multivariate gait observations and did not use GPPS as an input
variable.

The relative change between sessions was computed for each condition
from the condition-level mean GPPS values:

\[
\Delta_{\lambda}(\%)
=
\frac{
\overline{GPPS}_{M2,\lambda}
-
\overline{GPPS}_{M1,\lambda}
}{
\overline{GPPS}_{M1,\lambda}
}
\times 100,
\]

where $\lambda$ denotes the occlusal observational probe. This relative change is a descriptive longitudinal summary and does not, by itself, support causal or population-level inference.

\subsubsection{Level 2: Conceptual dynamical organization}
The system is modeled conceptually as a dynamical system:
\[
\frac{dX}{dt} = F(X,\lambda,\theta)
\]
where:
\[
\lambda = \mathrm{VDO}
\]
is treated as a constraint parameter associated with variations in system dynamics. This formulation is used as a conceptual framework and does not imply that the vector field is formally estimated.

\subsubsection{Level 3: Exploratory Multivariate Embedding}

Let

\[
D_{M1}=\{x_i\}_{i=1}^{N},
\qquad
x_i\in\mathbb{R}^{p},
\]

denote the M1 multivariate gait dataset, where each $x_i$ corresponds to one repeated gait observation and $N=270$.

A two-dimensional representation was defined as

\[
z_i=\Phi(x_i),
\qquad
\Phi:\mathbb{R}^{p}\rightarrow\mathbb{R}^{2},
\]

where $\Phi$ denotes the UMAP transformation \cite{mcinnes2018}.

Numerical variables were processed using a common pipeline across all six M1 observational probes. The six files were concatenated, yielding exactly \(N=270\) repeated gait observations. Fifty-five numerical biomechanical variables were retained. Empty technical variables, categorical identifiers, absolute identifiers, timestamps, and non-informative fields were excluded.

Missing numerical values were imputed using variable-wise medians estimated from the combined M1 analytical matrix. No IQR-based outlier replacement was applied in this Level~3 reconstruction. Each retained variable was standardized across the complete set of 270 M1 observations before dimensionality reduction.

The two-dimensional embedding was computed using UMAP with the following fixed parameters:

\[
n_{\mathrm{neighbors}}=15,
\qquad
\mathrm{min\_dist}=0.25,
\qquad
\mathrm{metric}=\mathrm{Euclidean},
\qquad
\mathrm{random\_state}=42.
\]

These parameters, together with the complete observation-level coordinates and the figure-generation script, were archived to ensure exact reproduction of the Level~3 analysis.

The resulting representation was used exclusively as an exploratory visualization of neighborhood relationships within the multivariate dataset. UMAP axes have no direct physical or physiological meaning. Absolute positions, distances, apparent clusters, and group boundaries should therefore not be interpreted as measurements of physiological state.

The Level~3 analysis did not use GPPS as an input variable. Accordingly, comparison between Level~1 and Level~3 concerns two distinct analytical representations: an aggregated scalar score and a model-dependent projection of multivariate gait observations.

\subsubsection{Level 4: Conceptual Extension}
A fourth analytical level is proposed as a purely conceptual extension describing potential relationships between system states under varying constraints.
This level is not implemented in the present study and is introduced solely as a conceptual extension of the proposed framework.
It does not rely on experimental data and does not correspond to a modeled or quantified dynamical process. Its role is to provide a theoretical perspective on how multivariate system configurations might be related within a broader configuration space. It does not assume that the Level~3 embedding identifies discrete or physiologically validated states.

\subsection{Separation Assessment}

The Level~3 representation was examined descriptively to determine whether the occlusal observational probes formed visually distinct groupings.

The interpretation focused on the balance between inter-condition displacement and within-condition dispersion. Because the observations were repeated measurements from one participant, the analysis was descriptive and was not interpreted as inferential evidence of population-level classes.

The revised M1 embedding showed substantial overlap among the six observational probes. Therefore, no condition was interpreted as defining an independently validated condition-specific cluster. In particular, any displacement between the OC2.5 and OC3 centroids was interpreted relative to their broad within-condition dispersion.

\subsection{Ethical Considerations}

The participant provided written informed consent for participation and publication of the data. The protocol was conducted in accordance with the principles of the Declaration of Helsinki.

\section{Results}

\subsection{Overview}

All occlusal observational probes produced measurable variations in spatiotemporal and plantar variables. These variations were not homogeneous and depended on both the probe and the measurement session. Two Level~1 representations are distinguished below. The first is the archived historical GPPS representation, retained for provenance. The second is the current reproducible GPPS, recalculated from the reconstructed analytical matrix after exclusion of the longitudinal term $I$ and correction of the CAPA weighting. The historical and current representations do not produce the same global ranking. This difference is treated as a result of weighting sensitivity rather than as evidence that either scalar representation uniquely identifies the underlying organization.

\subsection{Level 1: Historical and Current Aggregated Performance}

The archived historical GPPS representation ranked OC2.5 first and OC3 second at both sessions. Its stable ordering was:

\[
OC2.5 > OC3 > ONL > OSL > OC3P > OBL.
\]

These historical values predated the present multi-level analysis and are retained for provenance. However, the historical computation included a longitudinal progression term $I$ and did not preserve the complete observation-level transformation record. It is therefore not used as the current reproducible Level~1 result.

GPPS was recalculated using the fully disclosed formulation described in Section~2.6.1. The longitudinal term $I$ was excluded, asymmetries were represented by their absolute magnitudes, $CoP$ and $CAPA$ were each assigned a weight of 0.15, and one common global normalization was applied to the complete M1--M2 analytical matrix.

\begin{table}[H]
\centering
\small
\renewcommand{\arraystretch}{1.2}
\caption{
Comparison between the archived historical GPPS representation and the current reproducible Level~1 score. Historical values are reported for provenance. Current values were recalculated from 624 observation-level records using the disclosed formula without the longitudinal term $I$.
}
\label{tab:gpps}

\begin{tabular}{l c c c c}
\hline
Condition & Historical M1 & Historical M2 & Current M1 & Current M2 \\
\hline
ONL   & 8.3 & 8.5 & 4.721 & 6.452 \\
OSL   & 7.9 & 7.9 & 5.800 & 7.150 \\
OBL   & 6.2 & 6.5 & 6.255 & 7.373 \\
OC2.5 & 9.1 & 9.4 & 6.104 & 7.113 \\
OC3   & 8.8 & 9.1 & 6.662 & 7.455 \\
OC3P  & 7.5 & 7.8 & 5.839 & 6.527 \\
\hline
\end{tabular}

\end{table}

Under the current reproducible representation, OC3 obtained a higher mean GPPS than OC2.5 at both sessions. Their absolute differences nevertheless remained limited:

\[
|GPPS_{M1,OC2.5}-GPPS_{M1,OC3}|=0.558,
\]

\[
|GPPS_{M2,OC2.5}-GPPS_{M2,OC3}|=0.342.
\]

Thus, the relative ordering of OC2.5 and OC3 was reversed by the current score architecture, while their Level~1 proximity persisted. The robust observation is therefore their limited scalar discriminability, not the superiority of either condition.

\begin{figure}[H]
  \centering
  \includegraphics[width=0.98\textwidth]{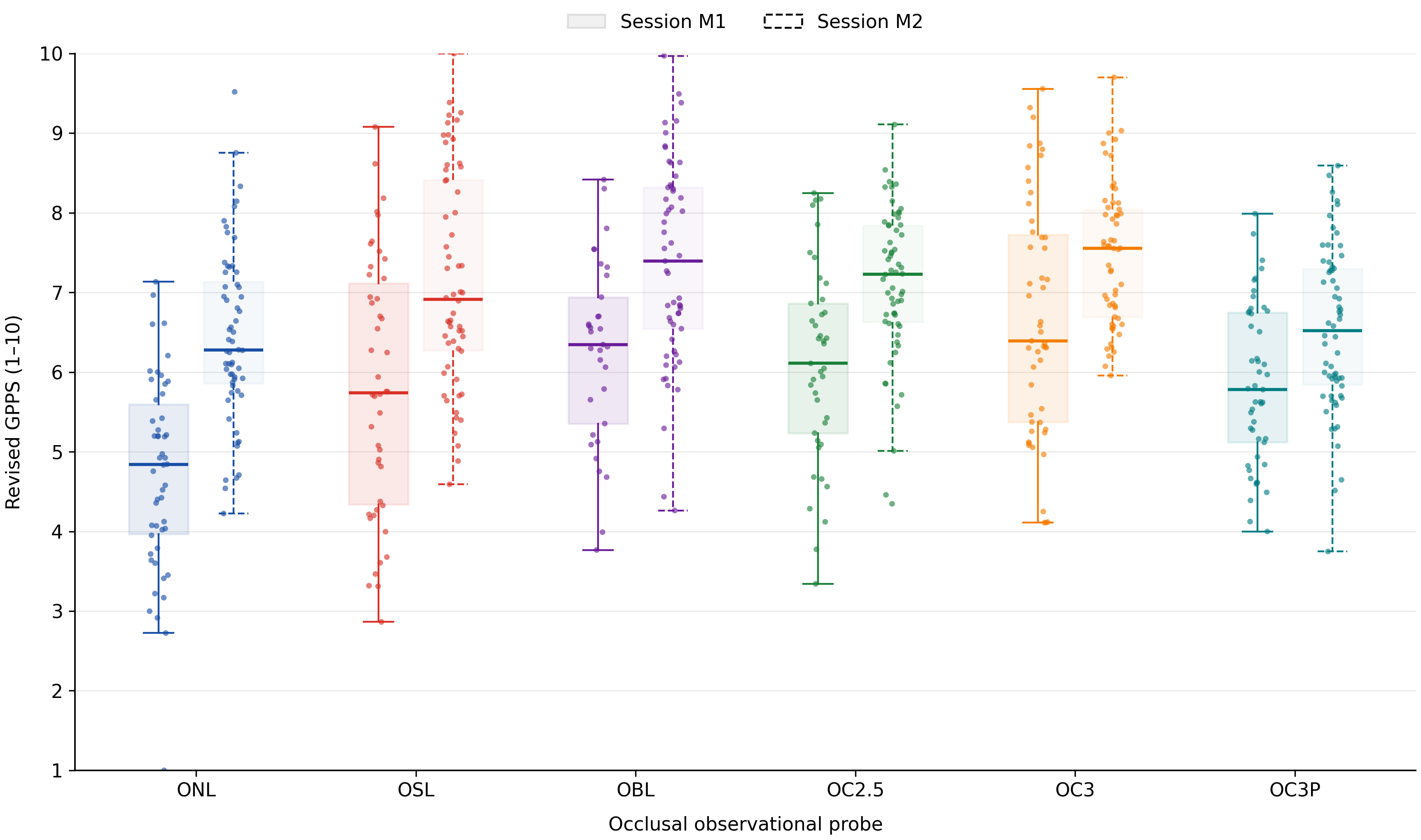}
  \caption{Reproducible distribution of current Level~1 GPPS values across occlusal observational probes. Each point represents one repeated gait observation from the single participant. Scores were recalculated from the globally normalized analytical matrix using the disclosed weighting scheme and excluding the longitudinal progression term $I$. Solid boxes represent M1 and dashed boxes represent M2. Boxes indicate the median and interquartile range. The observations should not be interpreted as independent subjects or population-level evidence. OC2.5 and OC3 retain partially overlapping scalar distributions, illustrating their limited discriminability at Level~1.
  }
  \label{fig:gpps_current}
  \end{figure}

Figure~\ref{fig:gpps_current} presents the observation-level distributions underlying the current reproducible scores.

The current score assigned relatively high values to OBL, although OBL was introduced clinically as an open-mouth perturbation probe. This result is not interpreted as evidence that OBL was physiologically favorable. Its scalar value was influenced by preserved velocity and cadence, by the orientation of plantar descriptors, and by the need to impute several unavailable derived variables in M1--OBL. OBL therefore provides an additional illustration of the limitation of scalar aggregation: preserved observable performance may coexist with a perturbing configuration or a non-equivalent compensatory organization.

\subsection{Level 2: Conceptual dynamical organization}

The dynamical analysis considers the evolution of the system in state space, with VDO considered as a constraint parameter:

\[
\frac{dX}{dt} = F(X,\lambda,\theta), \quad \lambda = \mathrm{VDO}
\]

Level~2 provides a conceptual state-space interpretation of temporal organization. No vector field, attractor, recurrence structure, or quantitative dynamical stability metric was estimated in the present study. Accordingly, Level~2 should be understood as an interpretive bridge between aggregated observation and exploratory multivariate representation, not as an experimentally validated dynamical model.

\subsection{Level 3: Exploratory Multivariate Representation at M1}

The revised Level~3 analysis was performed on the 270 observation-level records obtained during M1. Projection of the standardized multivariate dataset into two dimensions did not reveal six clearly separated condition-specific clusters.

\begin{figure}[H]
  \centering
  \includegraphics[width=\textwidth]{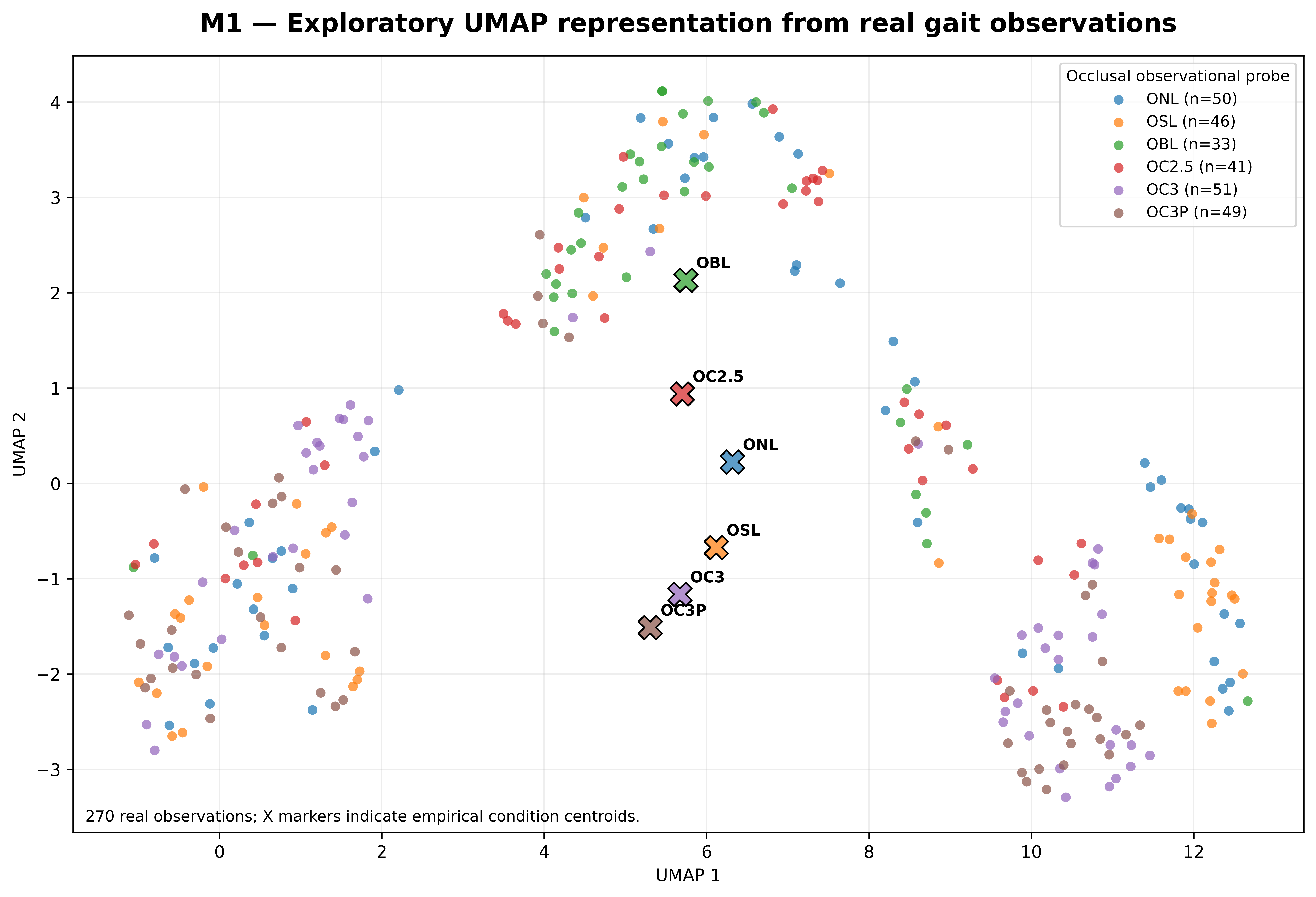}
  \caption{Exploratory UMAP representation of the 270 multivariate gait observations recorded during M1. Each point represents one repeated gait observation from the single participant and is colored according to the occlusal observational probe. The embedding was computed from 55 standardized numerical biomechanical variables using UMAP with \(n_{\mathrm{neighbors}}=15\), \(\mathrm{min\_dist}=0.25\), Euclidean metric, and \(\mathrm{random\_state}=42\). The six probes show substantial overlap, and no independently validated condition-specific clusters are observed. OC2.5 and OC3 display a limited displacement of their centroid positions but retain broad observation-level overlap. UMAP coordinates are model-dependent and have no direct physiological or metric interpretation.}
  \label{fig:level3_m1}
\end{figure}

As shown in Figure~\ref{fig:level3_m1}, observations from ONL, OSL, OBL, OC2.5, OC3, and OC3P occupied broadly overlapping regions of the selected UMAP representation. The within-condition dispersion was large relative to most inter-condition displacements. The embedding should therefore be interpreted as a continuous and overlapping multivariate organization rather than as a partition into discrete occlusal states.

OC2.5 and OC3 exhibited a limited displacement of their central positions within the selected representation. However, their observation-level distributions remained substantially overlapping. The M1 data therefore do not support the claim that OC2.5 and OC3 occupy clearly separated condition-specific regions within the selected representation.

The principal Level~3 result is consequently negative but informative: the occlusal observational probe did not uniquely determine the position of an observation within the low-dimensional multivariate representation.

\begin{figure}[H]
\centering
\includegraphics[width=0.82\textwidth]{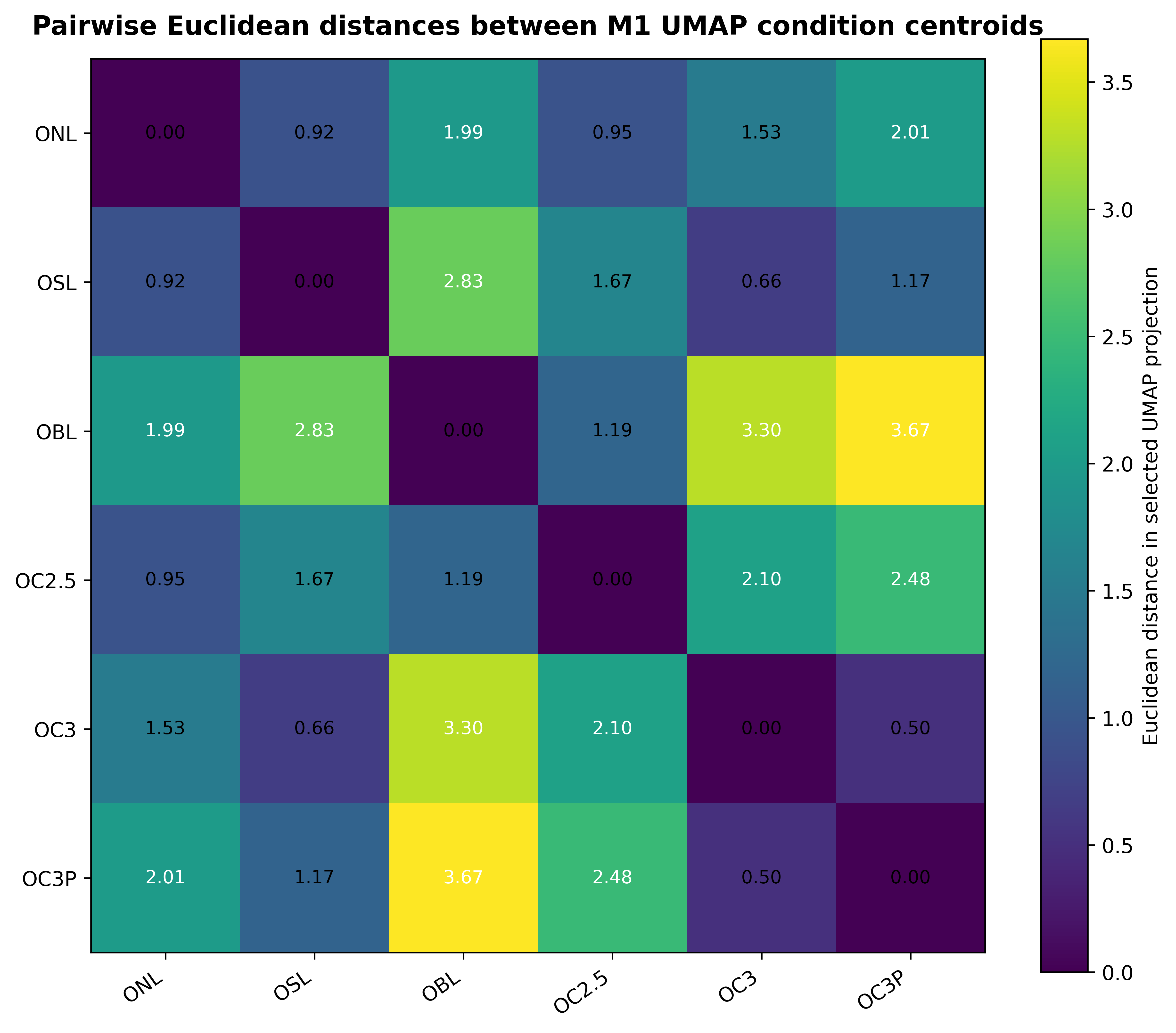}
\caption{
Pairwise Euclidean distances between the six condition centroids calculated directly from the same 270 observation-level UMAP coordinates displayed in Figure~\ref{fig:level3_m1}. These values summarize relative centroid displacement within this specific two-dimensional projection and should not be interpreted as physiological distances or as evidence of discrete condition-specific states. Centroid separation must be considered jointly with the substantial within-condition dispersion shown in Figure~\ref{fig:level3_m1}.
}
\label{fig:level3_centroids}
\end{figure}

Figures~\ref{fig:level3_m1} and~\ref{fig:level3_centroids} were generated within a single reproducible computational pipeline. The six condition centroids were calculated directly from the 270 observation-level UMAP coordinates, and the pairwise Euclidean distance matrix was then computed from those same centroid coordinates. Consequently, the point-level representation, centroid positions, and centroid-distance matrix are numerically consistent by construction.

Figure~\ref{fig:level3_centroids} summarizes the relative displacement of condition centroids within the selected projection. These centroid distances do not demonstrate cluster separation because they do not account independently for the broad observation-level dispersion. Their role is descriptive and complementary to the point-level representation.

\subsection{Non-Identifiability Across Representational Levels}

The revised analysis identifies two forms of non-identifiability. First, the relative ranking of OC2.5 and OC3 depends on the construction of the Level~1 composite score. Second, the Level~3 representation does not assign either condition to a compact and independently separated region.

Thus, neither the aggregated score nor the selected multivariate embedding provides a unique identification of an occlusal-condition-specific system state. Non-identifiability should therefore be understood here as the absence of a one-to-one correspondence between the experimental probe, the scalar performance score, and the position of an observation within the selected low-dimensional representation.

\subsection{Temporal Evolution}

Between M1 and M2, the current reproducible mean GPPS increased descriptively for all six probes. The magnitude of this change ranged from approximately 11.8\% to 36.7\%. These differences cannot be attributed to a specific causal factor, because the sensorimotor intervention, disease evolution, repeated measurement, fixed probe order, and other contextual elements were not isolated. The percentages are descriptive properties of the present score transformation and should not be interpreted as clinical treatment effects.

\section{Discussion}

\subsection{Main Finding}

The principal finding is not the identification of distinct physiological states associated with the six occlusal probes. The revised M1 analysis showed substantial overlap among all conditions in the selected UMAP representation.

The robust observation is instead a representational non-equivalence. Level~1 reduces each observation to an aggregated scalar score, whereas Level~3 preserves a broader set of multivariate relationships. However, the additional information retained by the multivariate representation did not organize the M1 observations into clearly separated condition-specific groups.

These results show that failure of a scalar score to discriminate between conditions does not imply that an unsupervised embedding will necessarily reveal discrete alternative organizations. In the present dataset, both levels display limited condition specificity, although they describe the observations in fundamentally different ways.

\subsection{Historical-to-Current Weighting Sensitivity}

The comparison between the archived historical and current reproducible GPPS representations shows that the relative ranking of occlusal probes depends on the architecture of the composite score. The historical representation ranked OC2.5 above OC3, whereas the current representation ranked OC3 above OC2.5. This inversion does not establish that one condition is clinically optimal or that one scalar representation is intrinsically correct. It demonstrates that ranking is conditional on variable orientation, normalization, weighting, and treatment of unavailable data.

This weighting-dependent non-identifiability provides the rationale for examining whether multivariate representations offer complementary descriptive information. In the revised M1 analysis, however, the higher-dimensional information did not produce a clear separation between OC2.5 and OC3.

\subsection{Implications for Biomechanics}

The present findings support the use of complementary analytical representations, but they also illustrate the limits of unsupervised visualization. A scalar score may conceal relationships among individual variables, whereas a low-dimensional embedding may preserve part of the local multivariate structure. Nevertheless, neither representation should be assumed to reveal the internal physiological organization of the system.

In this single-case M1 dataset, the six observational probes did not form discrete multivariate states. The biomechanical response appears better represented as an overlapping and variable organization than as a set of sharply separated condition-specific regimes.

\subsection{Interpretation of VDO as a Constraint Parameter}

Within this framework, VDO is not interpreted as a causal determinant but as a constraint parameter associated with variations in system organization. This interpretation aligns with constraint-based models in biological systems, where multiple functional states may coexist under similar observable performance conditions.

\subsection{Contribution of the Multivariate Representation}

The Level~3 analysis was based on a multivariate numerical matrix rather than on GPPS values. It therefore retained relationships among variables that were removed by scalar aggregation. Its contribution is representational rather than diagnostic: it provides a compact visualization of local similarities and differences among observations.

The absence of clearly separated condition-specific clusters is itself an important result. It indicates that the multivariate gait observations were not uniquely organized by the occlusal probe at M1. The selected embedding therefore does not demonstrate distinct internal physiological states, and its coordinates should not be interpreted as direct measures of system organization.

\subsection{Clinical and Conceptual Implications}

The historical Level~1 assessment favored OC2.5, whereas the current reproducible Level~1 score assigned a slightly higher mean value to OC3. The revised Level~3 analysis did not resolve this ambiguity because OC2.5 and OC3 remained broadly overlapping in the selected multivariate representation.

The study therefore does not identify an optimal occlusal configuration. More generally, it shows that neither a heuristic scalar ranking nor an unsupervised two-dimensional projection is sufficient, in isolation, to support clinical selection of an occlusal condition.

Clinical interpretation would require convergence among repeated measurements, clinically meaningful biomechanical variables, longitudinal representational behavior, patient-specific outcomes, and independently validated analytical criteria.

\subsection{Toward a Conceptual Level of State Relationships}

A fourth analytical level may be proposed as a theoretical representation of relationships among system configurations over time. However, the present Level~3 analysis did not identify discrete experimental states that could be used as empirically established nodes within such a model.

Level~4 therefore remains a prospective conceptual extension. It is not derived from the M1 embedding, does not represent observed transitions, and should not be interpreted as a causal, dynamical, predictive, or clinically validated model.

\subsection{Limitations}

This study has several limitations. First, it is based on a single-case design, which precludes generalization beyond the observed individual configuration. The recorded rows are repeated gait observations and not independent participants. Second, Parkinson's disease characterizes the functional context but is not used as an explanatory variable. Third, the fixed, non-randomized probe order confounds occlusal configuration with within-session position and possible adaptation, learning, or fatigue. Fourth, the uncontrolled sophrology-oriented sensorimotor intervention represents a longitudinal contextual factor that cannot be isolated.

Fifth, the archived historical GPPS values were not fully reproducible at the observation level and are retained only for provenance. Sixth, the current GPPS is a heuristic composite whose ranking remains sensitive to variable orientation, normalization, weighting, and imputation.

Seventh, the revised Level~3 analysis was restricted to M1 and therefore does not establish longitudinal evolution within a shared multivariate representation. Eighth, UMAP is a nonlinear and model-dependent dimensionality-reduction method whose output depends on preprocessing and parameterization. Ninth, the substantial overlap among conditions indicates that the observational probes did not define independently separated clusters in the selected representation. Tenth, centroid displacement within the embedding does not establish physiological separation and must be interpreted relative to within-condition dispersion.

Finally, the conceptual Level~2 and Level~4 formulations were not experimentally estimated as dynamical models. The analytical framework remains exploratory and does not support causal, predictive, mechanistic, diagnostic, or prescriptive inference.

\subsection{Perspectives}

Future work should reproduce the complete Level~3 pipeline using both M1 and M2 with a fixed preprocessing procedure, archived UMAP parameters, and explicit joint-embedding or out-of-sample projection rules.

Larger cohorts will be required to determine whether the overlap observed in this single participant reflects genuine within-person variability, insufficient condition specificity, or limited statistical power.

Future analyses should also compare UMAP with linear and nonlinear alternatives, including principal component analysis, diffusion maps, and autoencoder-based representations. Quantitative evaluation should include neighborhood preservation, within-condition dispersion, centroid uncertainty, permutation-based separation tests, and validation on independently acquired recordings.

Longitudinal analysis should prioritize the reproducibility and evolution of individual system representations over time rather than the visual appearance of isolated clusters.

\section{Conclusion}

This exploratory single-case analysis shows that interpretation of observable performance depends on the chosen analytical representation. The historical and current Level~1 scores produced different relative rankings of OC2.5 and OC3, demonstrating the sensitivity of scalar conclusions to weighting and normalization choices.

The revised M1 Level~3 analysis did not reveal clearly separated condition-specific clusters. The six occlusal observational probes occupied broadly overlapping regions of the selected UMAP representation, and OC2.5 and OC3 remained substantially overlapping despite a limited displacement of their central positions.

The central finding is therefore not that similar scalar performance conceals clearly distinct validated physiological states. Rather, neither the aggregated score nor the selected low-dimensional embedding provides a unique identification of an occlusal-condition-specific system state.

The principal contribution of this study is therefore methodological rather than physiological. It identifies representational non-identifiability as a legitimate scientific result that motivates the search for alternative analytical representations rather than stronger claims about the underlying biological system.

These results support a cautious multi-level framework in which scalar performance, exploratory multivariate representation, and longitudinal representational evolution are treated as distinct analytical properties. They do not establish causal occlusal effects, distinct physiological states, clinical thresholds, or an optimal vertical dimension.

\section*{Data Availability}

The raw clinical gait recordings are not publicly distributed because they derive from a single clinical participant. The anonymized derived analytical matrix, complete current Level~1 calculation, observation-level scores, condition-level summaries, preprocessing documentation, imputation audit, normalization bounds, and integrity hashes may be made available by the corresponding author upon reasonable request.

The derived M1 Level~3 analytical matrix, preprocessing record, UMAP configuration, observation-level coordinates, condition-centroid summary, and figure-generation script may also be made available. The archived Level~3 reproducibility package includes the complete 270-row UMAP coordinate file, the six condition-centroid coordinates, the derived centroid-distance matrix, the retained 55-variable analytical matrix, the imputation record, the fixed UMAP configuration, and the figure-generation script. Figures~\ref{fig:level3_m1} and~\ref{fig:level3_centroids} can therefore be reproduced from the archived numerical outputs without manual reconstruction. The historical condition-level GPPS values reported in Table~\ref{tab:gpps} predate the present multi-level analysis and are retained only for provenance.

\section*{Declaration of Generative AI Assistance}

Generative artificial intelligence tools, including ChatGPT (OpenAI), were used to assist with the creation of schematic illustrations and the graphical preparation of selected figures, as well as with translation of the manuscript into English. These tools are not authors or co-authors of this work. The human authors retain full responsibility for the accuracy and integrity of the figures, translations, scientific content, data, analyses, interpretations, and conclusions presented in the article.

\end{document}